\documentclass{article}
\usepackage{ijcai23}

\usepackage{times}
\usepackage{soul}
\usepackage{url}
\usepackage{multirow}
\usepackage[hidelinks]{hyperref}
\usepackage[utf8]{inputenc}
\usepackage[small]{caption}
\usepackage{graphicx}
\usepackage{amsthm}
\usepackage{booktabs}
\usepackage{algorithm}
\usepackage{algorithmic}
\usepackage[switch]{lineno}
\usepackage{amsmath,amssymb,amsfonts}
\usepackage{color}
\usepackage{subfigure}

\usepackage{titlesec}

\newcommand{\ie}{\textit{i.e.}}
\newcommand{\wrt}{\textit{w.r.t.\,}}

\title{\centering{\fontsize{14}{16}\textbf{$\mathcal{G}^2Pxy$: Generative Open-Set Node Classification on Graphs with Proxy Unknowns}}}

\author{\fontsize{12}{14}
\textbf{Qin Zhang$^1$},
\textbf{Zelin Shi$^1$},
\textbf{Xiaolin Zhang\footnote{Corresponding author. }},
\textbf{Xiaojun Chen$^1$},
\textbf{Philippe Fournier-Viger$^1$}, \textnormal{and \textbf{Shirui Pan$^2$}}
\affiliations
$^1$Big Data Institute, College of Computer Science and Software Engineering, Shenzhen University, China\\
$^2$School of Information and Communication Technology, Griffith University, Queensland, Australia
\emails
\{qinzhang, shizelin2021@email., xjchen@, philfv@\}szu.edu.cn,
solli.zhang@gmail.com,
s.pan@griffith.edu.au
}

\begin{document}

\maketitle

\begin{abstract}
Node classification is the task of predicting the labels of unlabeled nodes in a graph. State-of-the-art methods based on graph neural networks achieve excellent performance when all labels are available during training. But in real-life, models are often applied on data with new classes, which can lead to massive misclassification and thus significantly degrade performance.
Hence, developing open-set classification methods is crucial to determine if a given sample belongs to a known class. 
Existing methods for open-set node classification generally use transductive learning with part or all of the features of real unseen class nodes to help with open-set classification. 
In this paper, we propose a novel generative open-set node classification method, \ie, $\mathcal{G}^2Pxy$, which follows a stricter inductive learning setting where no information about unknown classes is available during training and validation.
Two kinds of proxy unknown nodes, inter-class unknown proxies and external unknown proxies are generated via \textsl{mixup} to efficiently anticipate the distribution of novel classes.
Using the generated proxies, a closed-set classifier can be transformed into an open-set one, by augmenting it with an extra proxy classifier.
Under the constraints of both cross entropy loss and complement entropy loss, $\mathcal{G}^2Pxy$ achieves superior effectiveness for unknown class detection and known class classification, which is validated by experiments on benchmark graph datasets. 
Moreover, $\mathcal{G}^2Pxy$ does not have specific requirement on the GNN architecture and shows good generalizations.
\end{abstract}

\section{Introduction}

Node classification is the task of assigning labels (categories) to unlabeled graph nodes. This task has many important  applications such as  traffic state prediction~\cite{zheng2020gman}, 
and completing user profiles in  social networks~\cite{wong2021standing}. Graph Neural Network (GNN) based methods~\cite{zhu2022shift,liu2021relative} achieve state-of-the-art performance for this problem. However, they generally assume that the full set of labels (\textit{known classes}) is available for training.
Unfortunately, this assumption does not hold in many practical scenarios because of unforeseen factors. For instance, a trained GNN may be applied on samples from a research social network where researchers may have published in new conferences that were unseen in training.
When facing samples of unknown classes, typical GNN methods~\cite{kipf2016semi-han10,velivckovic2018graph,hamilton2017inductive-han6} will incorrectly label nodes using known classes, which can considerably degrade performance.
Hence, building models that can classify samples from known classes and reject samples from unknown classes is a key issue, which is known as the \textsl{open set classification problem}.

Recent techniques that have been designed to address the open-set node classification problem~\cite{wu2021openwgl} presume that a model will be more confident in its predictions for closed-set samples than for open-set samples. Relying on this assumption, a threshold is applied on a model's confidence scores to separate unknown samples from known ones.  
Though this approach is intuitive~\cite{hendrycks2016baseline-han7-holder10}, it often fails because deep neural networks tend to be over-confident in input samples and may output high confidence scores even for an unknown input~\cite{hein2019relu-holder9,guo2017calibration-holder8,bendale2016towards-holder3-han2,scheirer2012toward-holder27}. 
Moreover, finding an optimal threshold to separate known from unknown classes is hard and time-consuming~\cite{perera2020generative-holder26,ge2017generative-holder6}. 

Another critical challenge for open-set classification is the lack of  open-set data. It is hard to collect open-set data due to the very large diversity. 
However, most of the prior solutions~\cite{wu2020openwgl} follow a transductive learning approach, i.e., suppose that part or all of the features of the real unseen class samples can be observed and used to help with open-set learning. This setting is not suitable for most real-world applications.
To alleviate this issue, several studies on image classification 
have attempted to estimate the distribution of unknown classes and calibrate a network's confidence. 
For example, G-OpenMax~\cite{ge2017generative-holder6} utilizes a GAN to generate unknown samples for training an image classifier. 
Neal et al.~\cite{neal2018open-holder22} introduced a counterfactual image generation method to produce image samples lying between decision boundaries through GANs. 
Perera et al.~\cite{perera2020generative-holder26} combined self-supervision and a generative model to
enhance the optimal closed-set features and to effectively identify open-set samples. 
P{\small ROSER}~\cite{zhou2021learning} learns placeholders to prepare for unknown classes by allocating placeholders for both data and classifier, which yields high disparity.

Motivated by the above studies, we propose a novel \textsl{\textbf{G}}enerative open-set node classification method on \textsl{\textbf{G}}raphs with \textsl{\textbf{Proxy}} unknowns ($\mathcal{G}^2Pxy$ for short).
Under the constraints of not being able to obtain open-set data, 
$\mathcal{G}^2Pxy$ first generates pseudo unknown class nodes (unknown proxies) via \textsl{mixup}~\cite{verma2019manifold-holder31,han2022g-gmixup}, to mimic the distribution of unknown classes using only the closed-set graph for learning. 
Two kinds of proxy unknown nodes, inter-class unknown proxies and external unknown proxies, are generated. 
Using the generated proxies,  a closed-set classifier can be transformed into an open-set one, by augmenting it with an extra proxy classifier. The whole model is optimized under the supervision of  cross entropy and tailored complement entropy loss~\cite{chen2018complement} together, and can then adaptively perform class-specific classification and detection during testing. 
Moreover, $\mathcal{G}^2Pxy$ can be adopted by different kinds of GNNs and shows good generalizations.
The contributions of this paper are summarized as follows:

\begin{itemize}
    \item A novel method $\mathcal{G}^2Pxy$ for open-set node classification is proposed with efficient unknown proxy generation and open-set classifier learning, under the constraint of both cross entropy and complement entropy loss. To the best of our knowledge, $\mathcal{G}^2Pxy$ is the first generative method for open-set graph learning. 
    \item No open-set data (samples of unknown classes or any side information of unknown classes) is required during training and validation. $\mathcal{G}^2Pxy$ has an explicit classifier for unknown and does not require tuning a threshold.
    \item $\mathcal{G}^2Pxy$ does not have specific requirement on the GNN architecture, and shows good generalizations. Using mixup as the generative method also ensures  the efficiency, due to the acceptable extra computation cost. 
    \item Experiments on benchmark graph datasets demonstrate the good performance of $\mathcal{G}^2Pxy$. 
\end{itemize}

\section{Problem Definition \& Preliminary}
This paper focuses on the node classification problem for a graph. 
Consider a \textsl{graph} denoted  as $G=(V,E,X)$, where $V=\{v_i | {i = 1,\ldots,N}\}$ is a set of $N$ nodes in the graph.
$E=\{e_{i,j} | i,j = 1,\ldots,N.$ $i\neq j \}$ is a set of  edges between pairs of nodes $v_i$ and $v_j$.
$X\in \mathbb{R}^{N\times d}$ denotes the feature matrix of  nodes, and $d$ is the dimension of node features. 
$x_i \in X$ indicates the feature vector associated with each node $v_i$. 
The topological structure of $G$ is represented as  an adjacency matrix $A \in \mathbb{R}^{N \times N}$, where $A_{i,j}=1$ if the nodes $v_i$ and $v_j$ are connected,~\ie, $\exists e_{i,j} \in E$, otherwise $A_{i,j}= 0$. 
$Y \in \mathbb{R}^{N \times C}$ is the label matrix of $G$, where $C$ is the already-known node classes. 
If node $v_i \in V$ is associated with a label $c$, $y_{i,c} = 1$, otherwise, $y_{i,c} = 0$.  

For a typical \textsl{closed-set node classification} problem, a GNN encoder $f_{\theta_g}$ takes node features $X$ and adjacency matrix $A$ as input, aggregates the neighborhood information and outputs representations.
Then, a classifier $f_{\theta_c}$ is used to classify the nodes into $C$ already-known classes.
The GNN encoder and the classifier are optimized to minimize the expected risk ~\cite{yu2017open-holder38} in Eq. \eqref{eq:closedrisk}, with the assumption that test data $\mathcal{D}_{te}$ and train data $\mathcal{D}_{tr}$ share the same feature space and label space, \ie,
\begin{equation}
    \label{eq:closedrisk}
    f^* = arg\min_{f\in \mathcal{H}} \mathbb{E}_{(x,y) \sim \mathcal{D}_{te}} \mathbb{I}(y\neq f(\theta_g, \theta_c; x,A))
\end{equation}
where $\mathcal{H}$ is the hypothesis space, $\mathbb{I}(\cdot)$ is the indicator function which outputs 1 if the expression holds and 0 otherwise. 
Generally, it can be optimized with cross-entropy to discriminate between known classes. 

In \textbf{open-set node classification problem}, given a graph $G=(V,E,X)$, $\mathcal{D}_{tr}=(X,Y)$ denotes the train nodes.
$\overline{\mathcal{D}}_{te}=(X_{te},Y_{te})$ is the test nodes, where  $X_{te}=S\cup U$, $Y_{te}= \{1,\ldots, C, C+1,\ldots\}$. $S$ is the set of nodes that belong to seen classes that already appeared in $\mathcal{D}_{tr}$ and $U$ is the set of nodes that do not belong to any seen class (\ie, unknown class nodes). Open-set node classification aims to learn a $(C+1)$-$class$ classifier $f_{\overline{\theta}_c}$ that $f(\theta_g,\overline{\theta}_c; X_{te},\overline{A}): \{X_{te},\overline{A}\}\rightarrow \overline{\mathcal{Y}}$, $\overline{\mathcal{Y}}= \{1, \ldots , C, unknown\}$, with the minimization of the expected risk \cite{yu2017open-holder38}:
\begin{equation}
    \label{eq:openrisk}
    \overline{f}^* = arg\min_{f\in \mathcal{H}} \mathbb{E}_{(x,y) \sim \mathcal{\overline{D}}_{te}} \mathbb{I}(y\neq f(\theta_g, \overline{\theta}_c; x,\overline{A}))
\end{equation}
where $\overline{A}$ is the adjacency matrix for $X_{te}$. 
The predicted class $unknown\in\overline{\mathcal{Y}}$ contains a group of novel categories, which may contain more than one class. 
Thus, the overall risk aims to classify known classes while also detecting the samples from unseen categories as class $unknown$. 

\textbf{From close-set to open-set classifier}, an intuitive way is thresholding~\cite{hendrycks2016baseline-han7-holder10}.
Taking the max output probability as confidence score, i.e., $conf=max_{c=1,\ldots,C} f_c(x,A)$, it assumes the model is more confident in closed-set instances than open-set ones. Then we can extend a closed-set classifier by 
\begin{equation} \label{eq:threshold}
    \hat{y}_i = \left\{
    \begin{aligned}
     & \arg\max_{c=1,\ldots C} f_c(x_i, A), \  conf > \tau \\
     & unknown, \ \ \ \ \ \ \ \ \ \ \ \   otherwise.
    \end{aligned}
    \right.
\end{equation}
where $\tau$ is a threshold. 
However, due to the overconfidence phenomena of deep neural networks, the output confidence of known and unknown is both high~\cite{bendale2016towards-holder3-han2}. 
As a result, tuning a threshold that well separates known from unknown is hard and time-consuming. 

Differently, the approach presented in this paper utilizes an additional proxy classifier to detect \textit{the unknown} class nodes, with well-designed generated unknown proxies to facilitate the model training. 
There are advantages of adding a class instead of optimizing a threshold  as in previous methods~\cite{shu2017doc-han22,wu2020openwgl,bendale2016towards-holder3-han2}, since it does not require real open-set samples in validation and also eliminates the difficulty of tuning the threshold.

\section{Methodology}
\begin{figure}[htbp]
\centering
\includegraphics[width=0.48\textwidth]{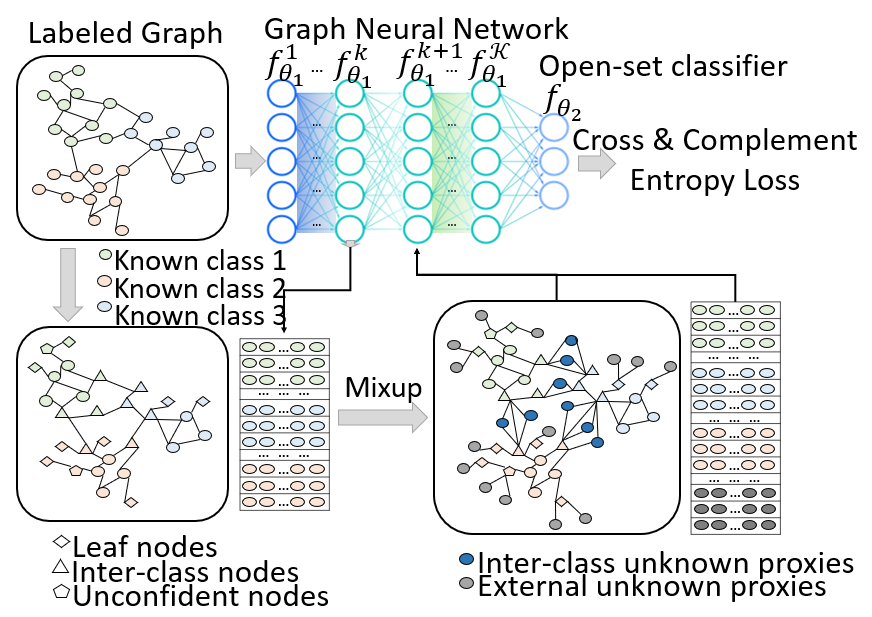} 
\caption{An overview of $\mathcal{G}^2Pxy$. Firstly, a graph with labeled nodes is input into the GNN model $f_{\theta_1}$. Inter-class and external unknown proxies are created in the middle of GNN (at the $k$-th layer). Then the generated proxies and the original training data are input into the rest of the neural network to learn an open-set classifier $f_{\theta_2}$, under the constraint of cross entropy and complement entropy loss. } 
\label{Fig:framework}
\end{figure}

Facing the difficulty of closed-set classifier calibration, we need proxy unknown nodes to foresee the distribution of unknown classes and prepare the open-set model for novel classes. 

The overall framework of $\mathcal{G}^2Pxy$ is introduced in Fig. \ref{Fig:framework}.
Specifically, a GNN encoder $f_{\theta_1}$ takes node features $X$ and adjacency matrix $A$ as input, aggregates the neighborhood information and outputs representation $h_{i}^k$ for each node $v_i$ in its $k$-th layer, $k=1,\ldots,\mathcal{K}$. $\mathcal{K}$ is the total number of layers. Thus, in the $k$-th layer, for node $v_i$, the GNN encoder aggregates neighbor information from the $k$-$1$ layer into a neighborhood representation:
\begin{equation}
    h_i^k = f^{k}(\theta_1; h_i^{k-1},h_{j}^{k-1}, j\in \mathcal{N}_i)
\end{equation}
where $h_i^k \in \mathbb{R}^{d_{k}}$ is the hidden representation at the $k$-th layer, $\mathcal{N}_i$ is the neighborhood of node $v_i$ and $h^0=X$.

Meanwhile, we mimic novel patterns by generating proxy unknown nodes through manifold mixup \cite{verma2019manifold-holder31,han2022g-gmixup}. 
Targeted node pairs are selected and mixed up at the middle layer of the GNN, and two kinds of proxies are created: inter-class unknown proxies that separate known classes from each other and external unknown proxies that separate known from unknown.
Generated proxies  and original known class nodes are input into the remaining layers together to learn discriminative representations. 
Then, a classifier $f_{\theta_2}$ is learned  to classify the nodes into $C+1$  classes, achieving open-set node classification.

To conclude, the proposed $\mathcal{G}^2Pxy$ method consists of two main modules: 1) a \textsl{proxy node generation} module to create proxy unknown class nodes; 2) an \textsl{open-set classifier learning} module 
to guarantee the classification of known classes and the rejection of the  unknown class. 
We introduce them in details. 

\subsection{Proxy Node Generation}

As out-of-distribution nodes are unavailable during training, we propose an effective way of generating proxy unknown nodes, which are designed to be located at the boundaries between different classes. 
These nodes are utilized in the learning process to supervise and tell apart unknown nodes from known ones.
Specifically, two different kinds of proxies are produced, inter-class unknown proxies and external unknown proxies. We adopt \textsl{manifold mixup}~\cite{verma2019manifold-holder31} as our generation method, due to its efficiency and good interpretability ~\cite{han2022g-gmixup}.

\subsubsection{Inter-Class Unknown Proxies}

For a well-trained classifier, 
features of nodes that belong to the same categories are close to each other, while those from different classes are distant.
Ideally, clear boundaries separate the different classes. 
Since nodes whose features are close to the boundaries are more likely to be less representative to their own classes, we decided to generate proxy unknowns using nodes near the boundary regions.
Thus, samples that belong to different classes but are close to each other are ideal node pairs for proxy generation, \ie,  node pairs where the two nodes are in different classes but have an edge between them. In details, given two connected nodes from different categories, i.e., $\{(x_i,y_i), (x_j,y_j)\}$, where $y_i \neq y_j$ and $a_{ij}=1$, their embeddings in the $k$-th layer are $h_i^k$ and $h_j^k$, 
obtained by giving the graph to the network $f_{\theta_1}^{1\sim k}$.
The inter-class unknown proxy $(\tilde{x}_i, \tilde{y}_i)$ are calculated by  Eq.~\eqref{eq:mixup-inter}.

\begin{equation} \label{eq:mixup-inter}
     \left\{
    \begin{aligned}
     & \tilde{x}_{i} = \alpha h_{i}^k(\theta_1; x_i, A) + \left ( 1 - \alpha \right ) h_{j}^k \left ( \theta_1; x_j, A \right ) \\
     & \tilde{y}_{i} =  C+1.
    \end{aligned}
    \right.
\end{equation}
where $\alpha \in \left [ 0, 1 \right ]$ is randomly sampled from a Beta distribution and the value of $\alpha$ is around $0.5$.
$\tilde{y}_{i}=C+1$ is the category index of $\tilde{x}_i$, indicating it belongs to the \textit{unknown}  class.
Two edges between $(x_i, \tilde{x}_i)$ and $(x_j, \tilde{x}_i)$ are also added into the graph. 
We denote the set of generated inter-class proxies samples as $X_{int}$. 
The purpose of $X_{int}$ is to simulate the distribution of unknown class nodes that exist between known classes. The mixed hidden representation $\tilde{x}_{i}$ is then passed to $f_{\theta_1}^{k+1\sim \mathcal{K}}$ 
to yield the final representation $\tilde{h}_i^{\mathcal{K}}=f_{\theta_1}^{\mathcal{K}}(\tilde{x}_i)$.

\subsubsection{External Unknown Proxies}

Besides the inter-class unknown proxies, external unknown proxies are further generated to mimic the unknown distributions at the periphery.
We use the peripheral nodes from each known class and their corresponding class center to generate the external unknown proxies. 

Specifically, the peripheral nodes are selected according to two rules.
\textit{Rule 1} is to select the leaf nodes of known classes, \ie, $x_i \in X_{tr}, \ s.t. \ \sum_{j}A_{i,j}=1$.
\textit{Rule 2} is to collect the nodes with low classification confidence, \ie,  the nodes which have the top $T$ least confident scores are selected in each known class. 
In this way, both structure-based and semantic-based peripheral nodes  are picked out. 
We record the set of peripheral nodes as $X_{per}$.

Moreover, the embedding of class centers in the $k$-th layer can be calculated  according to the ground-truth annotations following Eq.~\eqref{eq:center}:
\begin{equation}
    h_{(c)}^k = 
    \frac{1}{|X^c|} \sum_{x_i \in X^c} h_{i}^k  (\theta_1; x_i, A  ), c =1,\ldots, C.
    \label{eq:center}
\end{equation}
where $X^c$ denotes the nodes annotated with the class label $c$.

Then, the manifold mixup is applied on peripheral nodes and  their corresponding  reversed class centers $\{(x_i,y_i),(x_{(y_i)},y_i), x_i \in X_{per}, y_i \in \{1,\ldots,C\}\}$ to obtain the external unknown proxies in the $k$-th layer, i.e.,
\begin{equation} \label{eq:mixup-exter}
     \left\{
    \begin{aligned}
     & \tilde{x}_{i} =  h_{i}^{k}(\theta_1; x_i, A) + \alpha  ( - h_{(y_i)}^k)  \\
     & \tilde{y}_{i} =  C + 1.
    \end{aligned}
    \right.
\end{equation}
where $\alpha > 0$ is a hyperparameter to control the distance between the generated samples and the existing known class samples.
Edges between $(x_i, \tilde{x}_i), x_i \in X_{per}$  are also added into the graph. 
We denote the set of generated external unknown proxies as $X_{ext}$.

\subsection{Open-Set Classifier Learning}
To handle the diverse compositions of known-unknown categories, it is necessary to extract invariant information from the  known and unknown classes.
Thus, the backbone GNN network $f_{\theta_1}$ is shared by the $C+1$ classes to learn class distributions and node representations. 

Further, to calibrate closed-set (known classes) classifier to open-set classifier, an extra category is added to the last classification layer for unknown prediction. 
Suppose that the weights of the closed-set classifier is $w_{close}\in \mathbf{R}^{d_{\mathcal{K}}\times C}$ where $d_{\mathcal{K}}$ is the dimension of the embeddings in the last layer of GNN, the open-set classifier is the combination of the closed-set classifier and  the proxy classifier, \ie, ${\theta_2} = [w_{close}, w_{proxy}]\in \mathbf{R}^{d_{\mathcal{K}}\times (C+1)}$, where $w_{proxy}\in \mathbf{R}^{d_{\mathcal{K}}\times 1}$ is the weights \wrt the proxy classifier.
Here, the bias term is omitted for simplicity. 
Then,  the integrated classifier is trained with the original known class samples and the generated proxy samples,~\ie, $\overline{\mathcal{D}}_{tr}=\mathcal{D}_{tr} \cup (X_{pxy},Y_{C+1})$, $X_{pxy}=X_{int} \cup X_{ext}$, with the cross entropy loss in Eq. \eqref{eq:loss1}:
\begin{equation}
    l_1 = \sum_{ (x_i, y_i ) \in {\mathcal{D}}_{tr}} l_{CrE}(\hat{y}_{i},y_i) + \lambda_1 \sum_{x_i \in {X_{pxy}}} l_{CrE}(\hat{y}_{i},C+1) 
    \label{eq:loss1}
\end{equation}
where $\hat{y}_{i} = softmax(f_{\theta_2}(f_{\theta_1}(x_i)))$ is the output vector of the open-set classifier for  the input node $x_i$.
$l_{CrE}$ denotes the cross entropy loss, \ie, $l_{CrE}(\hat{y}_i,y_i)= - y_i\log\hat{y}_i$. 

Though the cross-entropy loss in Eq. \eqref{eq:loss1}  exploits mostly the information from the ground-truth class for maximizing data likelihood, the influence from complement classes (i.e., incorrect classes) has been largely ignored~\cite{chen2018complement}. 
Thus, we further adopt complement entropy \cite{chen2018complement} 
to complement the softmax cross entropy by neutralizing the effects of complement classes. 

The \textsl{complement entropy loss} $L_{CoE}$ is defined as the average of sample-wise entropies over complement classes, \ie,
\begin{equation}
    \label{eq:coeloss}
    l_{CoE}  = - \sum_{c=1,c\neq y_i}^{C+1} \frac{\hat{y}_{i,c}}{1-\hat{y}_{i,y_i}} \log \frac{\hat{y}_{i,c}}{1-\hat{y}_{i,y_i}}, \forall (x_i,y_i) \in \overline{D}_{tr}
\end{equation}
where $\hat{y}_{i,c}$ denotes the predicted probability of node $x_i$ belonging to class $c$.

For the needs of the generative open-set classification method $\mathcal{G}^2Pxy$, we further modify the above complement entropy according to the data distribution so that it better fits the task. 
Specifically, for generated unknown proxies,  we minimize their original complement entropy, \ie, minimize the average of entropy over the $C$ known classes, since the generated proxies could be adjacent to any of the known class. 
For the known class nodes, they are mostly adjacent to the generated proxies, \ie, the $C+1$ class, thus we push the proxy classifier to output the second-largest probability for nodes from known classes. According to this consideration, we propose a \textsl{tailored complement entropy loss} for $\mathcal{G}^2Pxy$ as follows:
\begin{equation}
    l_2  = \sum_{ (x_i, y_i ) \in {\mathcal{D}}_{tr}} l_{CrE}(\hat{y}_i/y_i, C+1 )  +  \sum_{x_i \in {X_{pxy}}} l_{CoE}(\hat{y}_{i},y_i)
    \label{eq:loss2}
\end{equation}
where $\hat{y}_i/y_i $ represents the prediction logits after removing the probability of its ground-truth label.
The first term helps to 
reduce the misclassification of known class nodes for other known classes. 
It is believed that with the help of loss in Eq. \eqref{eq:loss2}, the open-world classifier $f_{\theta_2}$ can classify nodes of known classes to their real categories, and use the proxy classifier to set an appropriate boundary to tell apart the known and the unknown. 
Finally, the total loss of $\mathcal{G}^2Pxy$ is a combination of cross entropy and the tailored complement entropy loss, \ie,
\begin{equation}
    l_{total} = l_1 + \lambda_2 l_2
    \label{overall loss}
\end{equation}
where $\lambda_2>0$ is the hyperparameter to balance these two losses. 

In the \textbf{test phase}, the well learned open-set classifier will be applied to a test node $x_i \in \overline{\mathcal{D}}_{te}$, and outputs $C+1$ probabilities $\hat{y}_i \in \mathbb{R}^{(C+1)\times 1}$. 
We choose the largest probability and give the related class label as prediction. Thus, $x_i$ will be detected as unknown class sample if the predicted label is $C+1$.

\section{Experiments}
Experiments were conducted to validate the performance of $\mathcal{G}^2Pxy$. 
These experiments include: \textsl{open-set node classification comparison}, \textsl{ablation study}, and \textsl{use case}.
Codes are available online\footnote{https://github.com/ejfomxhue2o3239djnwkk/G2Pxy}.

\paragraph{\textbf{Datasets.}}
Experiments to evaluate the performance for open-set node classification  were mainly carried out on four benchmark graph datasets~\cite{wu2020openwgl,zhu2022shift}, namely Cora\footnote{https://graphsandnetworks.com/the-cora-dataset/}, Citeseer\footnote{https://networkrepository.com/citeseer.php}, DBLP\footnote{https://dblp.uni-trier.de/xml/},
PubMed\footnote{https://pubmed.ncbi.nlm.nih.gov/download/},
which are widely used citation network datasets. 
Statistics are presented in Table \ref{tab:dataset}. 

\begin{table}[H]
\centering
\begin{tabular}{l|rrrr} 
\hline
Dataset  & Nodes & Edges &  Features &  Labels \\ \hline
Cora     & 2708       & 5429       & 1433          & 7           \\
Citeseer & 3312       & 4732       & 3703          & 6           \\
DBLP     & 17716      & 105734     & 1639          & 4           \\ 
PubMed   & 19717      & 44325      & 500           & 3           \\ \hline

\end{tabular}
\caption{Statistics of the experimental datasets.}
\label{tab:dataset}
\end{table}

\paragraph{Metrics.}

Accuracy and Macro-F1 are used for performance evaluation. 

\paragraph{Implementation Details.}
Generally, $\mathcal{G}^2Pxy$ adopts GCN~\cite{kipf2016semi-han10} as the backbone neural network for experimental evaluation unless otherwise specified. 
The GCN is configured with two hidden GCN layers in the dimensions of 512 and 128, followed by an additional multilayer perceptron layer of size 64.   
 $\mathcal{G}^2Pxy$ is implemented with PyTorch and the networks are optimized using stochastic gradient descent with a learning rate of $1e^{-3}$.
The balance parameters $\lambda_1$ and $\lambda_2$ are chosen by a grid search in the interval from $10^{-2}$ to $10^{2}$ with a step size of $10^{1}$. 

The baseline methods  are evaluated according to the instructions reported in the original papers with the same parameter configuration unless otherwise specified,  and  the best results are selected. 
For each experiment, the baselines and the proposed method were applied using the same training, validation, and testing datasets. 
The hyperparameters were tuned to get the best performance on the validation set.

All the experiments were conducted on a workstation equipped with an Intel(R) Xeon(R) Gold 6226R CPU and an Nvidia A100.

\paragraph{Test Settings.}
Compared to existing studies focusing on open-set  classification~\cite{wu2021openwgl,tang2021open,wang2020open}, the evaluation framework is extended. 
Two kinds of open-set classification evaluations were conducted to consider short or long distances between known classes and unknown classes, \ie, near open-set classification and far open-set classification. 
Besides, a discussion is provided of different settings in terms of the availability of side information of unknown class samples during training and validation, \ie, the inductive learning setting and the transductive learning setting. 

Specifically, in the near open-set classification experiment, for each dataset, the data of several classes were held out as the unknown classes for testing and 
 the remaining classes were considered as the known classes.  $70\%$ of the known class nodes were sampled for training, $10\%$ for validation and $20\%$ for testing. 
In the far open-set classification experiment, nodes from other datasets were used as unknown class samples for testing, instead of the nodes from the same dataset used in training. 
In experiments with inductive learning setting, there is not any information about the real unknown class (such as the features $x_i$ or side information of unknown classes) used during training or evaluation, while under the transductive learning setting, the whole graph (including unlabeled real unknown class nodes) are input during model training or validation. 

\paragraph{Baselines.}
The following methods are used as the baselines for the open-set classification comparison.
\begin{itemize}
    \item {} GCN\_soft: GCN~\cite{kipf2016semi-han10} is adopted for graph learning, where a softmax layer is used as the final output layer. It has no capability of recognizing the unknown classes.
    \item {} GCN\_sig: This baseline replaces the  softmax layer of GCN with multiple 1-vs-rest of sigmoids as the final output layer, which still does not have the capability of rejecting the unknown classes.
    \item {} 
    GCN\_soft\_$\tau$: Based on GCN\_soft, a  probability threshold  chosen in $\{0.1, 0.2,\ldots,0.9\}$ is used for the classification of each class . 
    \item {}
    GCN\_sig\_$\tau$: Similar to GCN\_soft\_$\tau$, a threshold is combined to GCN\_sig.
   \item {} GCN\_Opmax: Openmax~\cite{bendale2016towards-holder3-han2} is an open-set recognition model based on ``activation vectors''.
   Extreme value distribution is used to calibrate softmax scores to Openmax scores, which are then used for open-set classification. 
 
   \item {} GCN\_DOC: DOC~\cite{shu2017doc-han22} is an open-world classification method for text classification. It uses multiple 1-vs-rest of sigmoids as the final output layer and defines an automatic threshold setting mechanism. 
    GCN is combined with DOC to allow comparison with the proposed model. 
    \item{} GCN\_PROSER: PROSER \cite{zhou2021learning} is a novel and highly effective open-set recognition method used in the field of image processing. We paired it with GCN for graph-based comparison.
    \item {} OpenWGL: 
    OpenWGL~\cite{wu2020openwgl} employs an uncertainty loss in the form of graph reconstruction loss on unlabeled data and using an adaptive threshold to detect the unknown class samples. 
\end{itemize}

\begin{table}[tb]
\centering
\setlength{\tabcolsep}{0.8mm}{
\begin{tabular}{l|cc|cc|cc|cc}
\hline
\multirow{2}{*}{Methods} & \multicolumn{2}{c|}{Cora}     & \multicolumn{2}{c|}{Citeseer} & \multicolumn{2}{c|}{DBLP}     & \multicolumn{2}{c}{Pubmed}   \\ \cline{2-9} 
                         & Acc           & F1            & Acc           & F1            & Acc           & F1            & Acc           & F1            \\ \hline
GCN\_soft                & 70.6          & 67.6          & 44.6          & 38.9          & 63.8          & 59.2          & 28.9          & 29.9          \\
GCN\_sig                 & 69.2          & 64.7          & 45.3          & 44.5          & 63.5          & 58.7          & 28.9          & 29.8          \\
GCN\_soft\_$\tau$             & 73.6          & 73.8          & 57.3          & 54.5          & 65.0          & 62.4          & 49.7          & 48.6          \\
GCN\_sig\_$\tau$              & 79.7          & 80.1          & 62.1          & 54.6          & 69.2          & 68.2          & 45.1         & 46.0          \\
GCN\_Opmax               & 74.6          & 75.1          & 56.2          & 54.5          & 67.2          & 67.2          & 49.1          & 48.7          \\
GCN\_DOC                 & 77.8          & 78.1          & 66.0          & 56.7          & 69.9          & 69.2          & 45.6          & 46.2          \\
GCN\_PROSER     & 83.2          & 83.7          & 73.7          & 63.6          & 71.7          & 72.6          & 71.0          & 58.4 \\
OpenWGL     & 78.1          & 78.9          & 64.1          & 60.8          & 71.4          & 72.2          & 65.3          & 63.4 \\
$\mathcal{G}^2Pxy$                       & \textbf{84.3} & \textbf{84.8} & \textbf{75.5} & \textbf{71.0} & \textbf{77.3} & \textbf{79.0} & \textbf{73.7} & \textbf{70.2}          \\ \hline
\end{tabular}}
\caption{Near open-set classification on four citation network datasets with one unknown class (u=1) in the \textsl{inductive learning setting}. The Numbers reported are all percentage (\%).}
\label{tab:near-ood-detection-inductive}
\end{table}

\begin{table}[tb]
\centering
\setlength{\tabcolsep}{0.8mm}{
\begin{tabular}{l|cc|cc|cc|cc}
\hline
\multirow{2}{*}{Methods} & \multicolumn{2}{c|}{Cora}     & \multicolumn{2}{c|}{Citeseer} & \multicolumn{2}{c|}{DBLP}     & \multicolumn{2}{c}{Pubmed}    \\ \cline{2-9} 
                         & Acc           & F1            & Acc           & F1            & Acc           & F1            & Acc           & F1            \\ \hline
GCN\_soft                & 70.8          & 68.2          & 44.7          & 38.9          & 62.9          & 57.0          & 29.2          & 29.7          \\
GCN\_sig                 & 68.8          & 64.5          & 44.6          & 40.1          & 63.4          & 59.2          & 29.0          & 29.5          \\
GCN\_soft\_t             & 78.1          & 78.9          & 67.3          & 57.0          & 67.3          & 67.7          & 68.9          & 27.2          \\
GCN\_sig\_t              & 78.3          & 78.5          & 65.4          & 55.3          & 71.4          & 71.5          & 69.0          & 27.2          \\
GCN\_Opmax               & 77.2          & 76.9          & 57.5          & 56.7          & 69.0    & 70.6          & 55.0          & 52.1          \\
GCN\_DOC                 & 77.3          & 77.9          & 65.1          & 55.3          & 71.7          & 72.0          & 68.4          & 34.2          \\
GCN\_PROSER     & 84.7          & 83.6          & 74.3          & 66.6          & 75.3          & 71.6          & 72.8          & 60.8 \\
OpenWGL                  & 83.3          & 83.5          & 70.0          & 65.4          & 74.3          & 74.2          & 71.2          & 68.0 \\
$\mathcal{G}^2Pxy$                        & \textbf{90.7} & \textbf{89.7} & \textbf{76.3} & \textbf{71.8} & \textbf{77.5} & \textbf{79.5} & \textbf{78.0} & \textbf{73.4} \\ \hline
\end{tabular}}
\caption{Near open-set classification on four citation network datasets with one unknown class (u=1) under in the \textsl{transductive learning setting}. Numbers reported are all percentage (\%).}
\label{tab:near-ood-detection-transductive}
\end{table}

\begin{table*}[!htb]
\centering
\setlength{\tabcolsep}{2mm}{
\begin{tabular}{l|cc|cc|cc|cc|ccc}
\hline
\multirow{2}{*}{Methods} & \multicolumn{2}{c|}{Cora}     & \multicolumn{2}{c|}{Citeseer} & \multicolumn{2}{c|}{DBLP}     & \multicolumn{2}{c|}{Pubmed}   & 
\multicolumn{3}{c}{Average} \\
\cline{2-12} 
                         & ind           & ood            & ind           & ood            & ind           & ood            & ind           & ood
                         & ind
                         & ood
                         & overall
                     \\ \hline
GCN\_soft                & \textbf{89.7}          & 00.0          & 71.8          & 00.0          & \textbf{92.9}          & 00.0         & \textbf{93.2}          & 00.0 & \textbf{86.9} & 00.0 & 52.0         \\
GCN\_sig                 & 87.8          & 00.0          & 73.0          & 00.0          & 92.5          & 00.0          & 93.0          & 00.0 & 86.6 & 00.0 & 51.7         \\
GCN\_soft\_$\tau$             & 87.2          & 23.3          & 61.4          & 50.5          & 85.6          & 19.2          & 70.4          & 40.3 & 76.2 & 33.3 & 61.4      \\
GCN\_sig\_$\tau$              & 85.6          & 57.9          & 67.1          & 53.8      & 79.9          & 45.8          & 78.1          & 30.2 & 77.7 & 47.0 & 64.0          \\
GCN\_Opmax               & 86.6          & 30.0          & 60.4          & 49.4          & 85.4          & 27.6          & 73.4          & 38.1 & 76.4 & 36.3 & 61.8
\\
GCN\_DOC                 & 84.0          & 54.9          & 67.7          & 63.2          & 79.8          & 48.1          & 78.8          & 30.7 & 77.6 & 49.2 & 64.8         \\
GCN\_PROSER     & 83.2          & 82.7          & 72.5          & 76.0          & 78.2          & 58.0          & 78.0          & \textbf{70.2} & 78.0 & 71.7 &74.9
\\
OpenWGL     & 83.4          & 58.6          & 66.4          & 59.9          & 76.6          & 60.0          & 87.9          & 55.2 & 78.6 & 58.4 &69.7
\\
$\mathcal{G}^2Pxy$                       & 83.8 & \textbf{86.5} & \textbf{73.6} & \textbf{78.7} & 85.1 & \textbf{60.3} & 87.6 & 
67.5 &
82.5 &
\textbf{73.3} &
\textbf{77.7}
\\ \hline
\end{tabular}}
\caption{Detailed classification accuracy in terms of known classes ( in-distribution, \textsl{ind}) and unknown classes (out-of-distribution, \textsl{ood}) for near open-set classification on four datasets with one unknown class under inductive learning. Numbers reported are all percentage (\%).}
\label{tab:near-ood-detection-inductive-detail}
\end{table*}

\subsection{Open-Set Node Classification Comparison}
Since real-world scenarios are complex, where seen and unseen differs in diverse tasks, we evaluate our model from two aspects: near open-set classification, and far open-set classification. 
\subsubsection{Near Open-Set Classification} 
Table \ref{tab:near-ood-detection-inductive} and \ref{tab:near-ood-detection-transductive} list the accuracy and macro-F1 scores of the methods for the near open-set node classification task, 
where the last class of each dataset is set as the unknown class (i.e. u=1)  and the remaining classes are used to train the model. 
The results in Table \ref{tab:near-ood-detection-inductive} were obtained under the inductive learning setting, where there is no side information about the real unknown class samples available during training or validation. 
The results in Table \ref{tab:near-ood-detection-transductive} are obtained under the transductive learning setting, \ie, the features of unknown class nodes are used to facilitate the model training or validation. 

It is observed from these results that $\mathcal{G}^2Pxy$  outperforms the baselines on the datasets under both the inductive learning and transductive learning settings. This shows that $\mathcal{G}^2Pxy$  can better differentiate between a known class and an unknown class, though they are similar to each other. 
Specifically, $\mathcal{G}^2Pxy$ achieves an average of 3.70\% improvement over the second-best method (GCN\_PROSER) in terms of accuracy, and an average of 9.60\% improvement in terms of F1 score  on the  four datasets, under the inductive learning setting.
Also, $\mathcal{G}^2Pxy$ achieves 5.01\%  and 11.25\% improvements on average in terms of accuracy and F1 on the four datasets  under the transductive learning setting, respectively. 
Furthermore, by comparing Table \ref{tab:near-ood-detection-inductive} and Table \ref{tab:near-ood-detection-transductive}, it is observed that the involving of  real unknown class information  during training or validation benefits the model, which is reasonable. 

Besides, we show the detailed classification accuracy in terms of known classes (i.e. in-distribution classification, \textsl{ind} for short) and unknown classes (i.e. out-of-distribution classification, \textsl{ood} for short) respectively, in Table \ref{tab:near-ood-detection-inductive-detail}. 
The experiment was conducted under the same setting of Table \ref{tab:near-ood-detection-inductive}.
It shows that in order to gain the ability of unknown class detection, there is a slight decrease in the performance of known class classification, i.e. from 86.9\% to 82.5\% on average, comparing $\mathcal{G}^2Pxy$ to closed-set classification method GCN\_soft. However, the unknown class detection accuracy is improved from 0\% to 73.3\% on average, which is remarkable.
Compared to other open-set classification methods, such as the globally second best method GCN\_PROSER, the average performance of unknown-class node detection is improved from 71.7\% to 73.3\%  while the performance of the known-class classification still gain a certain increase from 78.0\% to 82.5\% on average.

\subsubsection{Far Open-Set Classification}
Following the protocol defined in~\cite{perera2020generative-holder26}, the models are trained and validated by all training and validation instances of the original dataset (in-distribution data, IND data). While for testing, instances from another dataset are augmented to the original test set as open-set classes (Out-of-distribution data, OOD data). 
Specifically, three experiments were designed: Cora\_Citeseer(\ie, Core as IND data and Citeseer as OOD data); Citeseer\_DBLP (Citeseer as IND, DBLP as OOD), and DBLP\_Pubmed (DBLP as IND, Pubmed as OOD), to evaluate the performance of the methods for the far open-set classification task. The known-unknown ratio is 1:1 by setting the number of test examples from the OOD dataset to the same number as in the IND dataset. 

The results for far open-set classification are shown in Table \ref{tab:farooddetection}. It is found that $\mathcal{G}^2Pxy$ can handle far out-of-distribution classes from diverse inputs and achieve a large performance improvement over state-of-the-art method GCN\_PROSER.
Surprisingly, the simple thresholding approach of  GCN\_soft\_$\tau$ achieves comparable performance with $\mathcal{G}^2Pxy$. This inspires us to combine the generative methods with discriminative methods for far open-set classification in  future work.  

\begin{table}[tb]
\centering
\setlength{\tabcolsep}{0.9mm}{
\begin{tabular}{l|cc|cc|cc}
\hline
\multirow{2}{*}{Methods} & \multicolumn{2}{c|}{\small Cora\_Citeseer}                            & \multicolumn{2}{c|}{\small Citeseer\_DBLP}                                        &\multicolumn{2}{c}{\small DBLP\_PubMed}  \\ \cline{2-7} 
                         & Acc       & F1       & Acc      & F1       & Acc & F1                \\ \hline

GCN\_soft            & 43.0 	        & 58.9 	        & 38.4 	        & 42.5 	             & 41.9            & 53.7             \\
GCN\_sig             & 41.6            & 57.5             & 36.3             & 42.1                  & 41.6            & 45.2           \\
GCN\_soft\_$\tau$         & 81.2 	        & 77.6 	        & 86.2 	        & 71.1                  & 85.0            & \textbf{75.6}   \\
GCN\_sig\_$\tau$          & 69.4            & 51.8             & 68.7             & 48.0                  & 79.8            & 69.1            \\
GCN{\small\_Opmax}           & 56.2 	        & 55.1 	        & 69.6 	        & 60.3 & 69.6            & 58.7            \\
GCN\_DOC             & 69.4 	        & 57.8 	        & 75.5 	        & 62.3 	             & 78.0            & 70.7             \\
GCN\_PROSER              & 78.5 	        & 79.1 	        & 81.5 	        & 66.4 	             & 78.6            & 69.0            \\
OpenWGL              & 80.6 	        & 76.7 	        & 44.6 	        & 11.9 	             & 84.6            & 70.7            \\
$\mathcal{G}^2Pxy$   & \textbf{81.3} 	& \textbf{80.5} 	& \textbf{87.5} 	& \textbf{74.4}         & \textbf{86.5}   & 72.3            \\
\hline 
\end{tabular}}
\caption{Accuracy and macro-F1 for far open-set classification on benchmark datasets. Numbers reported are all percentage (\%).}
\label{tab:farooddetection}
\end{table}

\subsection{Ablation Study}

\begin{table}[tb]
\centering
\setlength{\tabcolsep}{1mm}{
\begin{tabular}{l|cc|cc|cc|cc}
\hline
\multirow{2}{*}{Methods} & \multicolumn{2}{c|}{Cora}     & \multicolumn{2}{c|}{Citeseer}     & \multicolumn{2}{c|}{DBLP}          & \multicolumn{2}{c}{Pubmed}\\ \cline{2-9} 
                        & \multicolumn{1}{c}{Acc}       & F1     & \multicolumn{1}{c}{Acc}    & F1    & \multicolumn{1}{c}{Acc}  & F1     & \multicolumn{1}{c}{Acc}  & F1             \\ \hline
$\mathcal{G}^2Pxy$\_$\tau$         & \multicolumn{1}{c}{79.6}          & 80.5          & \multicolumn{1}{c}{71.4}          & 62.8          & \multicolumn{1}{c}{67.1}        & 67.1     &\multicolumn{1}{c}{70.0}    & 56.5  \\
$\mathcal{G}^2Pxy\_$int            & \multicolumn{1}{c}{84.2}          & 84.7          & \multicolumn{1}{c}{75.3}          & 70.8          & \multicolumn{1}{c}{75.3}        & 76.9     &\multicolumn{1}{c}{73.4}    & 68.7 \\
$\mathcal{G}^2Pxy\_$ext            & \multicolumn{1}{c}{84.1}          & 84.6          & \multicolumn{1}{c}{75.4}          & 70.9          & \multicolumn{1}{c}{75.5}        &74.8      & \multicolumn{1}{c}{71.4}   & 66.9 \\
$\mathcal{G}^2Pxy\neg${\small CoE} & 84.2                              & 84.7          &    75.2                           & 69.0         & 76.5                           &  77.7    & 70.1                       &47.3  \\

$\mathcal{G}^2Pxy$                & \multicolumn{1}{c}{\textbf{84.3}} & \textbf{84.8} & \multicolumn{1}{c}{\textbf{75.5}} & \textbf{71.0} & \multicolumn{1}{c}{\textbf{77.3}}     & \textbf{79.0}   &\multicolumn{1}{c}{\textbf{73.7}}  & \textbf{70.2} \\ \hline
\end{tabular}}
\caption{Accuracy and macro-F1 scores of $\mathcal{G}^2Pxy$ and its variants. Numbers reported are all percentage (\%).}
\label{tab:ablation}
\end{table}

We compare variants of $\mathcal{G}^2Pxy$ with respect to the generative strategy to demonstrate its effect.
The following $\mathcal{G}^2Pxy$ variants are designed for comparison.
\begin{itemize}
    \item {} $\mathcal{G}^2Pxy\_ \tau $: In this variant, generative strategies are removed, so the proxy unknown class is not involved in training. Instead, a threshold $\tau$ is set to separate the known and the unknown, which is chosen in $\{0.1,0.2,\ldots,0.9\}$. 
    \item {}
    $\mathcal{G}^2Pxy\_$int: A variant of $\mathcal{G}^2Pxy$ with only inter-class unknown proxies. 
    \item {}
    $\mathcal{G}^2Pxy\_$ext:
    A variant of $\mathcal{G}^2Pxy$ with only external unknown proxies. 
    \item {}
    $\mathcal{G}^2Pxy\neg$CoE:    A variant of $\mathcal{G}^2Pxy$ that the complement entropy loss (Eq. \eqref{eq:loss2}) is removed. 

\end{itemize}

Table \ref{tab:ablation} illustrates the results of these four variants and the proposed method. It can be seen that both the generated inter-class unknown proxies and the external unknown proxies matter. The proposed $\mathcal{G}^2Pxy$ method achieves significant improvements  over $\mathcal{G}^2Pxy\_\tau$  without unknown proxies, and also outperforms variants that only use a single type of unknown proxies. 
Meanwhile, the comparison between $\mathcal{G}^2Pxy$ and $\mathcal{G}^2Pxy\neg$CoE illustrates that the complement entropy loss is also indispensable. 

\subsection{Use Case}

The proposed model does not have specific requirement on the GNN architecture for classification. Hence, when unknown classes are expected in testing, the proposed  proxy node generation mechanism can be integrated and jointly optimized with convergence guarantee. The generated proxies can help find potential unknown class distributions and facilitate the rejection of  unknown class samples.

Table \ref{tab:backbone} compares the performance of the proposed  $\mathcal{G}^2Pxy$ with other open-set classification baselines, with different GNN architectures, including GCN, GAT and GraphSage. 
The results confirm the effectiveness and generalization ability of  $\mathcal{G}^2Pxy$  for open-set node classification. 

\begin{table}[htb]
\centering
\setlength{\tabcolsep}{0.5mm}{
\begin{tabular}{l|cc|cc|cc|cc}
\hline
\multirow{2}{*}{Methods}        & \multicolumn{2}{c|}{Cora}                                                   & \multicolumn{2}{c|}{Citeseer}                                              & \multicolumn{2}{c|}{Dblp}                                           &\multicolumn{2}{c}{PubMed}            \\ \cline{2-9} 
                                & \multicolumn{1}{c}{Acc}             & \multicolumn{1}{c|}{F1}             & \multicolumn{1}{c}{Acc}            & \multicolumn{1}{c|}{F1}              & \multicolumn{1}{c}{Acc}             & \multicolumn{1}{c|}{F1}           & \multicolumn{1}{c}{Acc}             & \multicolumn{1}{c}{F1}                 \\ \hline
{\small GCN\_$\tau$}                       & \multicolumn{1}{c}{ 73.6}           & \multicolumn{1}{c|}{ 73.8}           & \multicolumn{1}{c}{ 57.3}          & \multicolumn{1}{c|}{ 54.5}           & \multicolumn{1}{c}{ 65.0}           & \multicolumn{1}{c|}{ 62.4}             &\multicolumn{1}{c}{ 49.7}       &\multicolumn{1}{c}{ 48.6} \\
{\small GCN\_DOC  }                        & \multicolumn{1}{c}{ 77.8}           & \multicolumn{1}{c|}{ 78.1}           & \multicolumn{1}{c}{ 66.0}          & \multicolumn{1}{c|}{ 56.7}           & \multicolumn{1}{c}{ 69.9}           & \multicolumn{1}{c|}{ 69.2}             &\multicolumn{1}{c}{ 45.6}       &\multicolumn{1}{c}{ 46.2} \\ 
{\small GCN\_Opmax  }                      & \multicolumn{1}{c}{ 74.6}           & \multicolumn{1}{c|}{ 75.1}           & \multicolumn{1}{c}{ 56.2}          & \multicolumn{1}{c|}{ 54.5}           & \multicolumn{1}{c}{ 67.2}           & \multicolumn{1}{c|}{ 67.2}              &\multicolumn{1}{c}{ 49.1}       &\multicolumn{1}{c}{ 48.7} \\ 
{\small GCN\_$\mathcal{G}^2Pxy$  }         & \multicolumn{1}{c}{\textbf{ 84.3}}  & \multicolumn{1}{c|}{\textbf{ 84.8}}  & \multicolumn{1}{c}{\textbf{ 75.5}} & \multicolumn{1}{c|}{\textbf{ 71.0}}  & \multicolumn{1}{c}{\textbf{ 77.3}}  & \multicolumn{1}{c|}{\textbf{ 79.0}}    &\multicolumn{1}{c}{\textbf{ 73.7}}       &\multicolumn{1}{c}{\textbf{ 70.2}} \\ \hline
{\small GAT\_$\tau$   }                 & \multicolumn{1}{c}{ 71.6}           & \multicolumn{1}{c|}{ 69.2}           & \multicolumn{1}{c}{ 58.9}          & \multicolumn{1}{c|}{ 51.1}           & \multicolumn{1}{c}{ 65.4}           & \multicolumn{1}{c|}{ 66.6}                   & \multicolumn{1}{c}{ 43.2}     & \multicolumn{1}{c}{ 43.7}\\
{\small GAT\_DOC     }                   & \multicolumn{1}{c}{ 71.1}           & \multicolumn{1}{c|}{ 72.6}           & \multicolumn{1}{c}{ 62.4}          & \multicolumn{1}{c|}{ 59.5}           & \multicolumn{1}{c}{ 64.2}           & \multicolumn{1}{c|}{ 61.8}        & \multicolumn{1}{c}{ 42.1}     & \multicolumn{1}{c}{ 42.9}      \\
{\small GAT\_Opmax}                  & \multicolumn{1}{c}{ 66.3}           & \multicolumn{1}{c|}{ 63.4}           & \multicolumn{1}{c}{ 48.6}          & \multicolumn{1}{c|}{ 48.9}           & \multicolumn{1}{c}{ 62.5}           & \multicolumn{1}{c|}{ 56.9}          & \multicolumn{1}{c}{ 48.6}     & \multicolumn{1}{c}{ 47.0}     \\
{\small GAT\_$\mathcal{G}^2Pxy$ }        & \multicolumn{1}{c}{\textbf{80.4}}  & \multicolumn{1}{c|}{\textbf{81.0}}  & \multicolumn{1}{c}{\textbf{75.2}} & \multicolumn{1}{c|}{\textbf{70.9}}  & \multicolumn{1}{c}{\textbf{ 72.9}}  & \multicolumn{1}{c|}{\textbf{ 73.7}}     &\multicolumn{1}{c}{\textbf{ 71.7}}   &\multicolumn{1}{c}{\textbf{ 47.0}}\\ \hline
{\footnotesize G\text{-}Sage\_$\tau$ }                & \multicolumn{1}{c}{ 72.7}           & \multicolumn{1}{c|}{ 72.9}           & \multicolumn{1}{c}{ 63.5}          & \multicolumn{1}{c|}{ 51.2}           & \multicolumn{1}{c}{ 64.3}           & \multicolumn{1}{c|}{ 64.0}             & \multicolumn{1}{c}{ 46.6}     & \multicolumn{1}{c}{ 46.9}\\
{\footnotesize  G\text{-}Sage{\scriptsize\_DOC}    }              & \multicolumn{1}{c}{ 76.0}           & \multicolumn{1}{c|}{ 75.4}           & \multicolumn{1}{c}{ 63.6}          & \multicolumn{1}{c|}{ 59.9}           & \multicolumn{1}{c}{ 68.9}           & \multicolumn{1}{c|}{ 72.2}            & \multicolumn{1}{c}{ 44.6}     & \multicolumn{1}{c}{ 45.7}   \\
{\footnotesize  G\text{-}Sage{\_Opmax}  }            & \multicolumn{1}{c}{ 71.1}           & \multicolumn{1}{c|}{ 70.6}           & \multicolumn{1}{c}{ 47.9}          & \multicolumn{1}{c|}{ 48.7}           & \multicolumn{1}{c}{ 62.3}           & \multicolumn{1}{c|}{ 56.9}          & \multicolumn{1}{c}{ 44.4}     & \multicolumn{1}{c}{ 45.1}      \\
{\footnotesize  G\text{-}Sage}{\scriptsize\_$\mathcal{G}^2Pxy$ }  & \multicolumn{1}{c}{\textbf{87.2}}  & \multicolumn{1}{c|}{\textbf{87.3}}  & \multicolumn{1}{c}{\textbf{78.6}} & \multicolumn{1}{c|}{\textbf{76.9}}  & \multicolumn{1}{c}{\textbf{ 74.4}}  & \multicolumn{1}{c|}{\textbf{ 74.7}}     &\multicolumn{1}{c}{\textbf{ 72.8}}  & \multicolumn{1}{c}{\textbf{ 64.9}}\\ \hline
\end{tabular}}
\caption{Accuracy and macro-F1 scores of open-set classification methods with different backbone neural network. Numbers reported are all percentage (\%).}
\label{tab:backbone}
\end{table}

\section{Conclusion}

This paper introduced a novel generative method for open-set node classification, named $\mathcal{G}^2Pxy$. 
Unlike prior methods, it does not require side-information about unknown classes for model training and validation.
Two kinds of proxy unknown nodes, inter-class unknown proxies and external unknown proxies are generated to foresee the distributions of unknown classes.
Under constraint of both cross entropy loss and complement entropy loss, $\mathcal{G}^2Pxy$ achieves superior effectiveness for unknown class detection and known class classification, which is validated by experiments on benchmark graph datasets. 
Moreover, $\mathcal{G}^2Pxy$ also has good generalization since it has no specific requirement on the GNN architecture. 
In future work, we will try to find an effective combination of the generative model and discriminative model for near and far open-set node classification. 

\section*{Acknowledgments}

This research was supported by National Key R\&D Program of China(2021YFB3301500),
National Natural Science Foundation of China (62206179, 92270122),
Guangdong Provincial Natural Science Foundation (2022A1515010129, 2023A1515012584), 
University stability support program of Shenzhen (20220811121315001), 
Shenzhen Research Foundation for Basic Research, China(JCYJ20210324093000002).


\bibliographystyle{named}
\bibliography{reference}
\end{document}